\documentclass[submission,copyright,creativecommons]{eptcs}
\ifpdf
  \usepackage{underscore}         % Only needed if you use pdflatex.
  \usepackage[T1]{fontenc}        % Recommended with pdflatex
\else
  \usepackage{breakurl}           % Not needed if you use pdflatex only.
\fi
\usepackage{iftex}
\usepackage[numbers]{natbib}
\usepackage{amsmath}
\usepackage{graphicx}
\usepackage{multirow}
\usepackage{makecell}
\usepackage{multirow}
\newcommand{\tablefont}{\small}
\usepackage{times}
\usepackage{booktabs}
\usepackage{soul}
\usepackage{amsfonts}
\usepackage{url}
\usepackage{graphicx}
\usepackage{booktabs}
\usepackage{algorithm}
\usepackage[switch]{lineno}
\usepackage{xspace}
\usepackage{mathtools}
\usepackage{algorithm}
\usepackage[noend]{algpseudocode}
\usepackage{idpz3syntax}
\usepackage{float}
\lstdefinestyle{example}{
  basicstyle=\tiny,
  keywordstyle=\color{blue},
  morekeywords= {Context,Question,Options},
  breaklines,
  frame = single,
  captionpos=t,
}
\newcommand{\fodot}[0]{FO($\cdot$)\xspace}

\usepackage[dvipsnames]{xcolor}

\begin{document}

%\title[VERUS-LM: Combining LLMs with Symbolic Reasoning]{VERUS-LM: a Versatile Framework for Combining LLMs with Symbolic Reasoning}
\title{VERUS-LM: a Versatile Framework for Combining LLMs with Symbolic Reasoning}

%\author{Benjamin Callewaert
%\institute{Dept. of Computer Science, De Nayer Campus, KU Leuven -- Leuven.AI -- Flanders-Make, DTAI-FET}
%\email{benjamin.callewaert@kuleuven.be}
%\and Simon Vandevelde
%\institute{Dept. of Computer Science, De Nayer Campus, KU Leuven -- Leuven.AI -- Flanders-Make, DTAI-FET}
%\email{s.vandevelde@kuleuven.be}
%\and Joost Vennekens
%\institute{Vrije Universiteit Brussel, Brussels, Belgium}
%\email{joost.vennekens@vub.be}}
\author{Benjamin Callewaert \qquad\qquad Simon Vandevelde
\institute{Dept. of Computer Science\\De Nayer Campus, KU Leuven, Belgium}
\institute{Leuven.AI -- KU Leuven Institute for AI}
\institute{Flanders Make -- DTAI-FET}
\email{\{benjamin.callewaert, s.vandevelde\}@kuleuven.be}
\and
Joost Vennekens
\institute{Vrije Universiteit Brussel, Brussels, Belgium}
\email{joost.vennekens@vub.be}
}

\def\titlerunning{VERUS-LM: a Versatile Framework for Combining LLMs with Symbolic Reasoning}
\def\authorrunning{B. Callewaert, S. Vandevelde \& J. Vennekens}
\maketitle

\begin{abstract}
A recent approach to neurosymbolic reasoning is to explicitly combine the strengths of large language models (LLMs) and symbolic solvers to tackle complex reasoning tasks. However, current approaches face significant limitations, including poor generalizability due to task-specific prompts, inefficiencies caused by the lack of separation between knowledge and queries, and restricted inferential capabilities. These shortcomings hinder their scalability and applicability across diverse domains.
In this paper, we introduce VERUS-LM, a novel framework designed to address these challenges. VERUS-LM employs a generic prompting mechanism, clearly separates domain knowledge from queries, and supports a wide range of different logical reasoning tasks. This framework enhances adaptability, reduces computational cost, and allows for richer forms of reasoning, such as optimization and constraint satisfaction.
We show that our approach succeeds in diverse reasoning on a novel dataset, markedly outperforming LLMs.
Additionally, our system achieves competitive results on common reasoning benchmarks when compared to similar state-of-the-art approaches, and significantly surpasses them on the difficult AR-LSAT dataset.
By pushing the boundaries of hybrid reasoning, VERUS-LM represents a significant step towards more versatile neurosymbolic AI systems. All code and datasets required to reproduce the results in this text are available online: \url{https://gitlab.com/EAVISE/bca/verus-lm}
% \textbf{Max 14p including refs!}
\end{abstract}

\section{Introduction}
Logical reasoning is an essential aspect of problem-solving, decision-making, and critical thinking, making it an important goal in Artificial Intelligence.
While recent large language models~(LLMs) such as GPT-4~\citep{openai} and Gemini~\citep{team2023gemini} have shown an impressive leap in reasoning-like capabilities through techniques like chain-of-thought prompting, they often lack the ability to guarantee faithful and transparent reasoning.
Indeed, at a fundamental level, LLMs operate as black-box probabilistic models, which makes it difficulty to ensure accurate and coherent reasoning steps.
As a consequence, whether they are capable of ``true reasoning'' is still subject of much debate~\citep{Bender2021,Wu2024,Shojaee2025}.

Symbolic inference engines, on the other hand, can derive conclusions that are provably correct and typically also explainable. However, these systems struggle to interpret ambiguous natural language input. This creates an opportunity for neurosymbolic methods to combine the strengths of LLMs with symbolic reasoning systems. 
Interestingly, this method loosely parallels Kahneman's theory on human cognition, in which reasoning is divided into fast and unconscious thinking (System 1) and slow and methodical thinking (System 2)~\citep{Kahneman}.
Such approaches have already demonstrated impressive improvements on common reasoning datasets~\citep{logiclm,LLMASPKR,linc}. 

%To address these limitations, hybrid neurosymbolic approaches such asLogic-LM \citep{logiclm}, Linc \citep{linc} and LLM-ASP \citep{LLM-ASP} integrate LLMs with symbolic solvers. An LLM formulates problems into symbolic representations, which are processed by a symbolic solver, ensuring faithful reasoning and transparent results. This approach has demonstrated impressive improvements on datasets like ProofWriter \citep{proofwriter} and FOLIO \citep{folio}.
%
%\textbf{Ergens tussen deze paragrafen moet nog een voorbeeld komen}

Despite these promising results, current state-of-the-art approaches have several limitations that hinder their applicability across diverse domains. One significant drawback is their heavy reliance on task-specific prompts to generate symbolic representations, via in-context learning. This makes these approaches less general and harder to adapt to new domains, since this may require manual prompt engineering. In real-world scenarios, where problems are highly variable, task-specific prompts can cause a severe bottleneck. Furthermore, they typically try to reduce every user query to the same logical reasoning task (or to one of only a very limited set of reasoning tasks). In many domains, users have widely different kinds of queries, which correspond naturally to equally wide range of logical reasoning tasks (e.g., optimization, generating solutions, deriving consequences, and generating explanations).
Therefore, there is a lack of \textit{versatility} in the types of queries they are able to answer, limiting their effectiveness in complex domains where multiple forms of reasoning are required.

%Another fundamental issue lies in the absence of a clear separation between knowledge representation and the questions posed to the system. In many neurosymbolic models, the entire domain must be reformulated each time a new query is introduced. This inefficiency restricts reusability and increases computational costs, creating impractical barriers for applications requiring frequent or evolving queries. Real-life systems often involve iterative questioning over the same foundational knowledge, and without this separation, the potential of symbolic reasoning remains underutilized.

%These shortcomings are evident in performance evaluations on the \textbf{AR-LSAT} dataset \citep{arlsat}, which consists of law school admission test questions that span various types of reasoning tasks. The dataset reflects real-world complexity by requiring systems to analyze scenarios with nuanced conditions and derive logical conclusions. Current approaches struggle to generalize across the dataset's breadth, highlighting the necessity for more robust and adaptable neurosymbolic methodologies.

In this paper, we propose an enhanced neurosymbolic framework that addresses these limitations titles VERUS-LM. It features a generic prompting pipeline, a semantic refinement step, a clear separation between declarative knowledge from questions, and support for a wide range of logical reasoning tasks. In this way, it is able to handle complex and dynamic environments in a robust and effective manner. 

%Our system has been evaluated on dedicated task-specific benchmark datasets, including ProofWriter \citep{proofwriter}, \textbf{PrOntoQA} \citep{prontoQA}, Folio \citep{folio} and 
%\textbf{LogicalDeduction} \citep{logicaldeduction}, showcasing competitive performance executing specific logical reasoning tasks. On top of that our system was tested on our own  dataset consisting of diverse logical questions about 5 varying domains and the complex, versatile AR-LSAT dataset \citep{arlsat}, on which it exceeds the current state-of-the-art.

%Next, KB Reasoning also serves as a user-friendly interface for the IDP reasoning engine, developed at KU Leuven. Unlike other interfaces, such as the Interactive Consultant or traditional API, which require users to understand the logical inference mechanisms of the IDP system, KB Reasoning offers a more accessible alternative. By allowing users to pose questions in \input{paper}natural language, KB Reasoning makes the IDP engine available to a broader audience.
\section{Related Work}

Improving the reasoning capabilities of LLMs has recently become a topic of great interest.
For the purpose of this paper, however, we will not discuss works on improving LLM reasoning with in-context prompting or by fine-tuning, but refer to~\citep{Huang2023,Plaat2024} for more information on those techniques.
Instead, we will focus on the neurosymbolic approach in which an LLM is coupled with an explicit symbolic reasoning engine. %, which offers a promising path to systematically improve reasoning by combining the benefits of both systems.

Current state-of-the-art systems all largely work in the same way. Given a reasoning problem, which typically consists of a description of some domain and a query that needs to be answered, they first let an LLM generate a formal representation of the problem and then pass this on to a reasoning engine to solve.
In this way, they combine the benefits of both kinds of AI systems: the LLM offers a user-friendly natural language interface, while the reasoning engine ensures correct reasoning.
This offers a promising path towards improving the reasoning skills of LLM-based systems.
We will now briefly go over some of these systems and touch on how they differ.

In~\citep{LLMASPKR}, the authors present a pipeline to automatically generate Answer Set Programming~(ASP)~representations~\citep{ASP}.
Their method uses the LLM three times: first to extract relevant constants, next to generate predicates, and finally to generate the ASP rules.
Using a logic puzzle dataset, they demonstrate a 71\% increase in accuracy compared to a baseline GPT-4.
The same authors also rely on ASP in the [LLM]+ASP framework~\citep{LLMASP}, where given a problem description, the LLM extracts a set of atomic facts.
These facts, coupled with pre-defined, hand-crafted ``knowledge modules'' containing domain-specific ASP rules, are fed to an ASP solver to find a solution.
They validate their approach on four ``story-based'' datasets, on which it outperforms the state-of-the-art.

% Prolog paper?
In the LINC system~\citep{linc}, an LLM is used to generate statements in First Order Logic (FOL), which are given to a FOL reasoner.
%the Prover9 reasoner~\citep{prover9}.
To mitigate formalization errors, they generate $K$ formalizations and make use of $K$-way majority voting to decide on the correct response.
They demonstrate that their approach is competitive on the FOLIO dataset~\citep{folio}, and achieves remarkable improvements on the ProofWriter dataset~\citep{proofwriter}.

% The final neurosymbolic system we discuss, Logic-LM~\citep{logiclm}, has two distinctive features.
Finally, though similar to the previous approaches, Logic-LM~\citep{logiclm} has two distinctive features.
First, instead of relying on a single formalization language, their system supports logic programs, FOL, constraint satisfaction problems and SAT encodings.
In this way, the system is able to handle a broader range of problem types.
Second, they introduce a \textit{self-refinement stage}, in which the LLM iteratively attempts to fix any syntax errors based on feedback from the solver.
They validate their approach on five datasets, on which they generally outperform GPT-4 using chain-of-thought as baseline.
%They validate their approach on five datasets\joost{Ik denk niet dat het nodig is om hier al die datasets op te nomen, namely ProofWriter, FOLIO, PrOntoQA~\citep{prontoQA}, LogicalDeduction~\citep{logicaldeduction} and AR-LSAT~\citep{arlsat}}, on which they generally outperform GPT-4 using chain-of-thought as baseline.

It is worth noting that most of the aforementioned methods rely heavily on dataset-specific examples, utilized for in-context learning, to guide the LLMs.
Furthermore, they only support a limited number of logical reasoning tasks, which limits their generalization capabilities to new datasets.
This is especially apparent on the AR-LSAT dataset~\citep{arlsat}, featuring logical reasoning questions from the Law School Admission Test, which is high in diversity and thus requires broad reasoning capabilities.
For instance, given knowledge on the types of CDs sold by a store, the questions can range from ``Which CDs are on sale?'' to ``Given $\phi$, which statements are true?'' and ``What are the minimum CDs required to be on sale if $\phi$?'', for some additional statement $\phi$.
Though it still improves on the baseline GPT-4, Logic-LM only achieves a 43.04\% accurary on this dataset due to its diversity.
%Due to this diversity, Logic-LM only achieves a 43.04\% accuracy on this dataset, though still an improvement over baseline GPT-4.

%\begin{lstlisting}[mathescape,style=example]
%Context: A music store carries exactly ten types of CDs, both new and used of each of jazz, opera, pop, rap, and soul. The store is having a sale on some of these types of CDs. The following conditions must apply: Used pop is on sale. [...] If either type of rap is on sale, then no soul is on sale.
%Question: If both types of jazz are on sale, then which one of the following is the minimum number of types of new CDs that could be included in the sale?
%Options: A) one B) two C) three D) four E) five
%\end{lstlisting}

Though different from the others, we also briefly discuss the \mbox{Symb}CoT framework~\citep{symbocot} for the sake of comparison.
Like the aforementioned works, SymbCoT internally translates a problem into FOL using in-context learning.
However, instead of using a separate reasoning engine, the LLM itself then performs logical reasoning on the generated statements in a step-by-step way similar to Chain-of-Thought, instructed by specific prompts that try to mimic the kinds of logical reasoning found in logical solvers. By not using a separate reasoning engine, the approach becomes more robust to syntax errors. 
Interestingly, this approach consistently outperforms standard Chain-of-Thought and Logic-LM on the PrOntoQA~\citep{prontoQA}, FOLIO and ProofWriter datasets.

\section{System Design}
\subsection{Requirements}
As discussed in the previous section, current state-of-the-art approaches typically focus on a single reasoning tasks.
However, more realistic use cases will likely require a more versatile approach.
% However, in a realistic use case, a more versatile approach is required.
%Imagine, for instance, a chat bot acting as a digital assistant in the field of financial lending.
%A person interacting with such an AI system will not only ask questions like ``Am I eligible for a loan?'', but will also ask other things such as ``Why am I (in)eligible?'' or ``If I wish to loan money for a house, what loan terms are possible based on my personal financial information?''.
Imagine, for instance, a chat bot acting as a digital assistant in the field of car insurance.
A person interacting with such an AI system will not only ask questions like ``Am I eligible for an insurance policy?'', but will also ask other things such as ``Why am I (in)eligible?'', ``Depending on my car type, what would my insurance premium?'' and ``What elements can I change to minimize my premium?''.

For a system to be capable of handling such use cases, there are two main requirements.
First, the system must support multiple modes of reasoning, which means that reasoning-specific LLM prompts will not work well (as evidenced by the results of state-of-the-art systems on the AR-LSAT dataset).
Second, the reasoning component of such a system must be capable of performing different kinds of reasoning, either using a single versatile engine or, similar to Logic-LM, using multiple ones. However, this latter approach has the downside that the domain knowledge needs to be formalized anew for each type of reasoning, increasing the computational cost and risk of errors. We therefore prefer the former approach of using a single reasoning engine that supports different forms of reasoning.

Given this, we have the following design requirements for VERUS-LM:
\begin{enumerate}
    \item \textit{Versatile}: it should support multiple \textit{useful} forms of reasoning (e.g., verification, explanation, optimisation, etc.)
    \item \textit{Knowledge reuse}: it should distinguish between ``domain knowledge'' and ``task to be performed'', allowing the domain knowledge to be reused for different tasks.
    \item \textit{Generic}: all aspects of the tool, including the prompting method, should be independent of the type of reasoning tasks or the kind of domain that is being considered. %\joost{Ik heb dit hier ``generic'' genoemd ipv ``problem-agnostic'', omdat die tweede term klinkt alsof het enkel onafhankelijk moet zijn van het probleem en niet ook van het probleemdomain.} \simon{Top!}
\end{enumerate}

% Misschien moeten we het zo kaderen: de huidige state of the art gebruikt reasoning engines om te redeneren, maar focust typisch op 1 taak en werkt niet in een interactieve context. Een voorbeeld van een reele situatie is een systeem om leningen te bepalen: een gebruiker wil niet enkel weten of ze een lening kunnen krijgen, maar ook waarom, welke lening voor hun de minste rente heeft, welke looptijden allemaal mogelijk zijn, enz. Met andere woorden: een systeem moet dus manieren ondersteunen om te redeneren. Daarbovenop merken we dat de meeste systemen een (soms zeer) domain-specifieke prompt meegeven om betere resultaten te halen, wat een groot verschil maakt (zoals aangetoond in de LSAT dataset, zie later). Daarom zijn onze vereisten als volgt:
% 
% \begin{enumerate}
%     \item Problem agnostic prompting mechanism, mimicking workflow of KE
%     \item Detecting inference + multiple inference support
%     \item Clear separation between Domain Knowledge and Question
% \end{enumerate}

\subsection{Reasoning Engine}

While the aforementioned requirements pertain to the general design of our system, they also constrain our selection of reasoning engine.
%Indeed, the domain of symbolic reasoning is large, not all reasoners are created equal: each has their own properties and emphases, making them more or less suited for our purposes.
% We will now briefly introduce our reasoning engine of choice, and motivate it.
To serve as the main reasoning core of VERUS-LM, we have selected the IDP-Z3 system~\citep{IDP-Z3}.
IDP-Z3 is a reasoning engine for \fodot, a rich extension of classical First-Order Logic with useful features such as types, aggregates, (inductive) definitions, and more. It explicitly supports the Knowledge Base Paradigm~\citep{KBP}, in which the same domain knowledge can be used for many different reasoning tasks. %, such as satisfiability checking, generating logical models, deriving consequences, and explaining consequences.  %In particular, it supports 
%is stored indmodeled independent of its use.
%Knowledge is stored declaratively in an \fodot Knowledge Base (KB), to which a broad range of inference tasks can be applied to derive relevant information.
%For instance, given a KB, IDP-Z3 can verify its satisfiability, generate possible solutions, derive consequences, explain why someting must/cannot be true, and more.

%By extending classical FOL with some additional concepts, the \fodot language is well-suited for general knowledge representation.
%Among others, \fodot adds 
%An \fodot KB is generally structured in three main components: the vocabulary, structure, and theory.
%We will briefly go over these blocks using the example given below, and refer to~\citep{IDP-Z3} for a more extensive overview.
We will briefly illustrate \fodot and the forms of reasoning supported by IDP-Z3 through a small example, referring to~\citep{IDP-Z3} for a more extensive overview.

As in normal FOL, a \emph{vocabulary} in \fodot consists of a set of symbols. Because \fodot is a typed logic, this includes types, in addition to predicate and function symbols. For instance, in the below example about car insurance, we use types \texttt{Customer} and \texttt{Car}.
%The function \texttt{interest\_rate} maps each \texttt{Duration} to the associated interest rate ($\in \mathbb{R}$), while the functions \texttt{age} maps each \texttt{Customer} to their age ($\in \mathbb{Z}$).
The function \texttt{risk\_factor} maps each \texttt{Car} to its associated risk factor ($\in \mathbb{R}$), while the function \texttt{age} maps each \texttt{Applicant} to their age~($\in \mathbb{Z}$).
A number of constants (= 0-ary functions) represent the car's value, its type, and the insurance premium. Finally, a unary predicate \texttt{applicant} represents which of the customers is requesting the insurance.
%A number of constants (= 0-ary functions) represent the amount of some specific loan, the duration of this loan, and the total amount due. Finally, a unary predicate \texttt{Borrower} represents which of the customers are taking out this particular loan.
Again following FOL, a \emph{structure} for a vocabulary provides an interpretation for some (not necessarily all) of the symbols in this vocabulary.
Finally, a \emph{theory} consists of a set of logical sentences.
In this simple example, we have two sentences: one states that every applicant must be an adult (L18), and one defines the calculation of the insurance premium (L19).
% \simon{Optional: split example in two columns for space}

%\begin{lstlisting}[language=IDP]
%vocabulary V {
%    type Duration := {5..25} <: Int
%    type Customer := {Ann, Brit} 
%    interest_rate: Duration -> Real
%    loan_amount, total: () -> Real
%    loan_duration: () -> Duration 
%    borrower: Customer -> Bool
%    age: Customer -> Int
%}
%structure S:V {
%    interest_rate := {5 -> 0.0221, ..., 25 -> 0.0249}.
%    age := { Ann -> 16, Brit -> 32}.
%}
%theory T:V {
%    !p in Customer: borrower(p) => age(p) >= 18.
%    total() = loan_amount()  * (1 + interest_rate(loan_duration()))**loan_duration().
%}
%\end{lstlisting}
%\begin{minipage}{0.45\textwidth}
%\begin{lstlisting}[language=IDP]
%vocabulary V {
% type Customer := {Ann, Brit}
% type Car := {Sedan,SUV,...}
% age: Applicant -> Int
% risk_factor: Car -> Real
% car_value, premium:  -> Real
% car_type:  -> Car
% applicant: Customer -> Bool
% premium: -> Real
%}
%\end{lstlisting}
%\end{minipage}
%\begin{minipage}{0.54\textwidth}
%\begin{lstlisting}[language=IDP,firstnumber=11]
%structure S:V {
% age := {Ann -> 16, Brit -> 32}.
% risk_factor := {Sedan->1.03,Truck->1.15, ...}.
%}
%theory T:V {
% !p in Customer: applicant(p) => age(p) >= 18.
% premium() = (car_value()/100) 
%            * risk_factor(car_type()). 
%}
%\end{lstlisting}
%\end{minipage}

\begin{lstlisting}[language=IDP]
vocabulary V {
 type Customer := {Ann, Brit}
 type Car := {Sedan,SUV,...}
 risk_factor: Car -> Real
 age: Applicant -> Int
 car_value: -> Real
 premium:  -> Real
 car_type:  -> Car
 premium: -> Real
 applicant: Customer -> Bool
}

structure S:V {
 age := {Ann -> 16, Brit -> 32}.
 risk_factor := {Sedan->1.03,Truck->1.15, ...}.
}

theory T:V {
 !p in Customer: applicant(p) => age(p) >= 18.
 premium() = (car_value()/100) * risk_factor(car_type()). 
}
\end{lstlisting}

As usual, we denote the value of a symbol $\sigma$ in a structure $S$ by $\sigma^S$ and extend this notation to terms and formulas (e.g., \texttt{age(Ann)}$^S = 16$).  Also as usual, a structure $S$ such that $\phi^S = \texttt{true}$ for all $\phi\in T$ is called a \emph{(logical) model} of $T$ and this is written as $S\models \phi$.

Given such a KB, IDP-Z3 can perform different forms of reasoning to answers different questions. 
For instance, IDP-Z3 can determine which customers are eligible for an insurance policy, and explain why.  Given the car type and car value, it can also calculate the insurance premium.
Or, it can ``optimize'' the premium by looking for the car type with the lowest risk factor.
Using these different reasoning tasks does not require any modifications to the KB itself, allowing for straightforward knowledge reuse.

These properties of IDP-Z3 make it a natural choice for our reasoning system.
Additionally, \fodot is expressive enough to model many problem domains, as supported by real-life use cases~\citep{Aerts2022,Deryck2019,Vandevelde2022,Vandevelde2024}.
Furthermore, as First-Order Logic is the subject of many articles and textbooks, the training corpora of LLMs will also contain enough information to allow them to be fluent in this formalism.

\subsection{VERUS-LM Architecture}
\begin{figure*}
    \centering\includegraphics[width=1\linewidth]{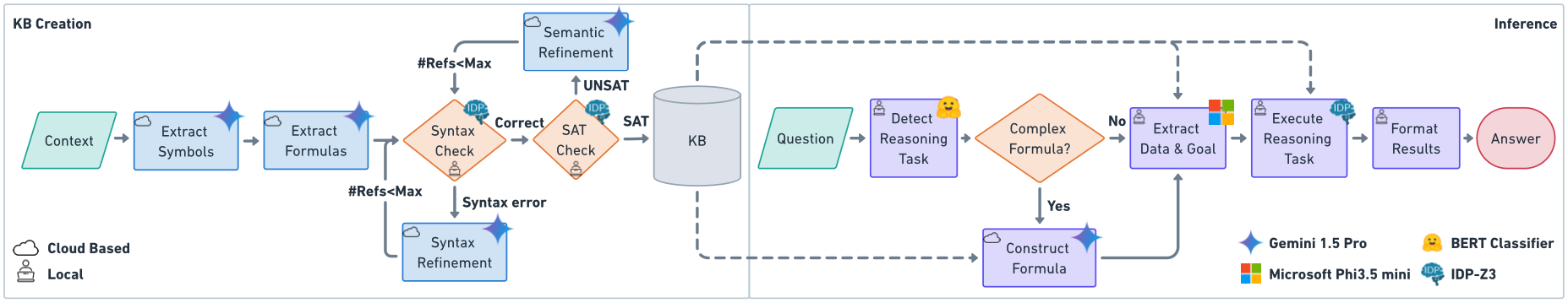}
    \caption{Flowchart depicting steps of VERUS-LM framework and the used system in each step}
    \label{fig:VERUS-LM}
\end{figure*}

The VERUS-LM framework integrates natural language processing and formal reasoning through a structured, two-phase process, as depicted in Figure~\ref{fig:VERUS-LM}. The first phase, Knowledge Base Creation, uses an LLM to translate domain knowledge into a symbolic \fodot specification. Note that, in this phase, we do not yet consider specific questions and instead create a reusable KB. Questions are only considered in the second phase, the Inference phase, where they are interpreted and then answered by means of an appropriate call to one of IDP-Z3's reasoning tasks. In other words, when given multiple questions about the same domain it is only formalized once, greatly reducing the computational cost. %This clear separation between the domain knowledge and question facilitates the reusability of KBs, effectively reducing the computational overhead of redefining the domain knowledge.

\section{Knowledge Base Creation}

%Although LLMs struggle with logical reasoning, they excel at interpreting natural language and  translating it into structured formats like mathematical equations \citep{LLM_math} or Python code \citep{LLM_python}.

The first phase of the VERUS-LM framework consists of transforming domain knowledge into a formal KB. %, inspired by LLM's capability of translating natural language to structured formats like equations~\citep{LLM_math} or Python code~\citep{LLM_Python}.
This translation is performed in three steps, which we will discuss in the following sections.

% The first phase of the VERUS-LM framework leverages this to transform domain knowledge into a formal \fodot Knowledge Base (KB).
%The translation is performed in three steps, where an LLM first defines the vocabulary before constructing a theory for it. %Urged by the fact that \fodot is a typed (multi-sorted) version of FOL, this is typically the approach Knowledge Engineers would follow. 
%Next, a self-refinement step validates the constructed KB and fixes it if necessary. %ensures that the defined KB is syntactically correct and satisfiable, mirroring the verification and validation steps typically done during Knowledge Acquisition. 
%We now elaborate on each of these steps individually.

\subsection{Symbol Extraction}
The LLM first identifies the relevant types, functions, and predicate symbols for the problem domain. We use a prompt that describes these three different kinds of symbols and how they are typically used. Important here is that we do not include examples from a specific dataset. Instead, our prompt just provides general instructions on how to identify and categorize the different concepts, and provides three general examples that do not appear in any of the datasets as an illustration.
The prompt also contains the instruction to annotate each symbol with its informal language meaning, so that this information is present in the next phases of the pipeline.
%LLM Each symbol is associated with an informal natural language meaning that reflects its role in the domain. % \joost{$\leftarrow$ Wat bedoel je hier juist mee? Wordt die informal language meaning ergens bijgehouden / ergens voor gebruikt?}
%\benjamin{Yes, die wordt bijgehouden zodat deze mee kan genomen worden naar de inference phase, dit is vooral belangrijk bij vb volgordes Book:(Nr) $\rightarrow$ BookOrder // Order form left to right, maar ook height:() $\rightarrow$ Int // Height in meters}

%Types represent disjoint subsets of the domain, providing a way to sort the domain elements. Predicates define relationships between elements of different types, while functions map elements from one type to another.

%A vocabulary is used to declare these symbols, representing the important concepts in the problem domain. 
%Together, these symbols form the \fodot vocabulary. %, unifying the building blocks of the domain.

One limitation of IDP-Z3 is that it does not support reasoning under the Open World Assumption, which is necessary for the FOLIO and Proofwriter datasets. Instead, IDP-Z3 always makes the Closed Word Assumption (CWA). In particular, it requires that the extension of all types (apart from built-in types $\mathbb{R}$ and $\mathbb{Z}$) is always fully enumerated. As a work-around, VERUS-LM simulates OWA reasoning by introducing an additional ``unknown'' domain element into each type.
Even though this is not theoretically correct, it persistently yields the correct results in practice, as shown in the experiments described further on.
% \simon{hier moeten we ook iets over zeggen}

%While this is not true OWA reasoning, it yields equivalent results in practical scenarios.
%\simon{Als we ergens ruimte te weinig hebben kunnen we evt het stukje over OWA omvormen naar een korte voetnoot, ofzo.}

\subsection{Formula Extraction}%\joost{Ik zou deze stap hernoemen, zowel in de figuur als in de titel van deze sectie als in de eerste zin. Zie Teams voor suggesties.}
During the formula extraction step, an LLM generates an \fodot theory that corresponds to the description of the problem domain. To guide this generation process, a brief overview of the \fodot grammar and clear instructions are presented, supported by two illustrative examples. These examples adopt a step-by-step approach, pairing each logical sentence with its corresponding natural language meaning for clarity. Additionally, the LLM is instructed to also consider implicit and commonsense knowledge, demonstrated in the examples. At the end of the prompt we add the previously generated vocabulary, along with the natural language description of the problem to formalize.
%, the vocabulary that was produced in the previous step is presented, along with the natural language description of the domain, which is to be translated to \fodot.

\subsection{Self-Refinement}
Although in our experience the LLM often produces correct logical sentences, errors still occur. To correct these, we introduce a self-refinement step that uses feedback from the reasoning engine to correct erroneous logical statements. This approach is inspired by recent work on imperative programming \citep{chen2023teaching,madaan2024self} and the Logic-LM framework \citep{logiclm}.  However, whereas these approaches only consider syntactic correctness, we introduce a novel second step to take semantic correctness into account as well, by means of a satisfiability check.
These refinements are part of an iterative process, where VERUS-LM will repeat them until no errors are found or a maximum of attempts is reached.

%\joost{Ik heb ``teaches LLMs to debug their generated logic through error feedback and few-shot demonstrations'' hier verwijderd, omdat dat klinkt alsof er retraining gebeurt van het LLM (``teaches LLMs to\ldots''), terwijl we dat niet doen.}

\paragraph{Syntax Refinement}
If the reasoning engine finds a syntax error, the LLM is instructed to correct its output, i.e., either the vocabulary and/or the theory it produces. To this end, the LLM is prompted with its previous erroneous output, the solver’s error message, and some examples of common syntax errors and their remedies.
%In most cases, syntax errors arise from incorrect predicate or function signatures. 
% Note that the IDP-Z3 solver is equipped with a code analysis tool, FOLINT \citep{folint} that provides detailed error descriptions.
Detailed error descriptions are provided by FOLINT~\citep{folint}, IDP-Z3's code analysis tool.

\paragraph{Semantic Refinement}
%\benjamin{Blijven we dit semantic refinement noemen of toch SAT refinement}
%\joost{Ik zou spreken van ``Semantic'' refinement ipv ``UNSAT refinement''}
Once the KB is syntactically correct, we use a satisfiability check to find additional semantic errors. It is reasonable to assume that the KB should be satisfiable, because, at this point, it contains only a formalization of the provided domain knowledge.
%, and not yet any information to do with a specific question that needs to be answered.
For instance, if we were solving a planning problem, the KB would contain a description of the actions that can be executed, but we would not yet  include a specific goal that the plan should achieve.
%As this KB does not yet contain any information w.r.t. the defined question, we can reasonably assume that the given domain knowledge is supposed to be satisfiable, i.e, that there exists at least one scenario which satisfies all constraints. 
If this KB is already inconsistent, then we take this as a sign that the LLM made a mistake in translating the domain knowledge and we instruct it to refine its previous output, based on an explanation of why this was unsatisfiable. IDP-Z3 can give such an explanation itself, by generating a ``minimal unsatisfiable subset'' of conflicting assignments and constraints~\citep{IDP-Z3}.%Here, it is useful that whenever IDP-Z3 encounters an inconsistency, it is able to generate an explanation for it, in the form of a ``minimal unsatisfiable subset'' \citep{IDP-Z3} i.e., a set of conflicting assignments and constraints, which we then include in the prompt. 

% Fixing the errors is an iterative process, where VERUS-LM will repeat syntax and semantic refinements until no errors are found or a maximum total refinements is reached.
%Note that in each refinement it is not only possible to refine the theory but also to change the vocabulary. %\joost{Heb ik correct geinterpreteerd wat je bedoelde met ``not only refine the constraints but also to alter the defined domain symbols''?}
%\benjamin{Ja, in iedere refinement is het dus idd mogelijk om constraints aan te passen maar ook de vocabulary aan te passen}
%\begin{algorithm}
%\small
%\caption{Translate Knowledge Base to IDP File}
%\begin{algorithmic}[1]
%\Procedure{TranslateKB}{$Problem$, $OWA$}
    %\State $symbols \gets extract\_symbols(Problem)$
    %\State $symbol\_idp \gets make\_idp(symbols, ``'', OWA)$
    %\State $i \gets 0$
    %\While{not $correct\_idp(symbol\_idp)$ \textbf{and} $i < 4$}
        %\State $symbols \gets refinement(problem, idp, syntax)$
        %\State $symbol\_idp \gets make\_idp(symbols, ``'', OWA)$
        %\State $i \gets i + 1$
    %\EndWhile
    %\State $constraints \gets extract\_constraints(problem, symbols)$
    %\State $j \gets 0$
    %\While{True \textbf{and} $j < 5$}
        %\State $idp \gets make\_idp\_file(symbols, constraints, OWA)$
        %\If{not $correct\_idp(idp)$}
            %\State $symbols, constraints$ $\gets$ $refine( idp, syntax)$
        %\ElsIf{not $sat\_idp(idp)$}
            %\State $symbols, constraints$ $\gets$ $refine( idp, sat)$
        %\Else
            %\State \textbf{break}
        %\EndIf
        %\State $j \gets j + 1$
    %\EndWhile
    %\State \Return $idp$
%\EndProcedure
%\end{algorithmic}
%\end{algorithm}

\section{Inference}
%\begin{figure*}[t]
%    \centering
%    \begin{minipage}[t]{0.49\linewidth}
%        \centering        \includegraphics[width=\linewidth]{img/normal-chatidp.png}
%        \\(a) KB Reasoning in normal mode
%    \end{minipage}%
%    \hfill
%    \begin{minipage}[t]{0.49\linewidth}
%        \centering
%        \includegraphics[width=\linewidth]{img/interactive-chatidp.png}
%        \\(b) KB Reasoning in interactive mode
%    \end{minipage}
%    \caption{Overview of steps executed by ChatIDP. (a) shows the flowchart of KB Reasoning in normal mode, while (b) illustrates KB Reasoning in interactive mode.}
%    \label{img:flowchart}
%\end{figure*}

Once the KB is constructed, the inference phase attempts to correctly answer the question.
To this end, VERUS-LM follows a series of steps, as illustrated on right side of Figure~\ref{fig:VERUS-LM}.
Given a question, the framework starts by identifying the intended reasoning task and extracting the relevant information, which are passed on to the reasoning engine. Once the latter has finished, its output is formatted back to natural language.

\subsection{Detect Reasoning Task}
%As listed in our design requirements, supporting different logical reasoning task is crucial for solving divers logical questions. The supported reasoning task of VERUS-LM, include seven built-in IDP reasoning tasks and a dedicated reasoning task, required for the FOLIO and Proofwriter dataset: 
%As listed in our design requirements, supporting different forms of logical reasoning  is crucial for solving divers logical questions.
First, VERUS-LM classifies the user's question into one of the following eight forms of reasoning. %supports seven of IDP-Z3's built-in reasoning tasks, and adds a custom reasoning task for logical entailment.  % Hi Benjamin, ik wou hier vermijden dat we zeggen dat VERUS-LM een reasoning task heeft voor twee specifieke datasets -- een reviewer met een slechte dag zou dat interpreteren als "probleem-specifiek" en weinig generaliseerbaar. Maar het is uiteraard wel generaliseerbaar. :-)
\begin{enumerate}
    \item \textbf{Model Generation/Expansion}: generate $n$ logical models of the theory $T$ (i.e., structures $S$ such that $ S\models T$). Possibly, the interpretation $\sigma^S$ that some of the symbols $\sigma$ should have in this model $S$ is already given (e.g., we might provide the value for constants \texttt{car\_value} and \texttt{car\_type} and ask the reasoner to complete the interpretation for the remaining symbols).
    \item \textbf{Satisfiability}: verify if at least one model $S$ exists for a given theory $T$. 
    \item \textbf{Optimization}: find the model $S$ of a given $T$ in which a given term $t$ reaches its minimal/maximal value $t^S$.
    \item \textbf{Propagation}: determine which atomic formulas are true / false in \emph{all} models $S$ of the given theory $T$.  
    \item \textbf{Explain}: explain why a given atomic formula is true / false in all models of $T$, or, in $T$ has no models, then explain the inconsistency.
    \item \textbf{Determine Range}:  determine the range of possible values for a given function term $f(\vec{t})$ given a theory $T$, i.e., the set of all values $v$ such that there exists at least one model $S$ of $T$ in which $f^S(\vec{t}^S) = v$. 
    \item \textbf{Relevance}: determine which symbols $\sigma$ are  relevant, in the sense that there exists a model $S$ of $T$ such that $S$ would no longer be a model if the value of $\sigma$ in $S$ were different.
    \item \textbf{Logical entailment}: verify whether a given statement $\phi$ is logically entailed by theory $T$.
\end{enumerate}

%\paragraph{Model Generation/Expansion} generate $n$ logical models of the given theory $T$ (i.e., structures $S$ such that $ S\models T$). Possibly, the interpretation $\sigma^S$ that some of the symbols $\sigma$ should have in this model $S$ is already given (e.g., we might provide the value for constants \texttt{loan\_duration} and \texttt{loan\_amount} and ask the engine to fill in the interpretation for the remaining symbols).
%\paragraph{Satisfiability}verify if at least one such model $S$ exists a given theory $T$. 
%\paragraph{Optimization} find the model $S$ of a given $T$ in which a given integer term $t$ reaches its minimal/maximal value $t^S$.
%\paragraph{Propagation} determine which atomic formulas are true / false in \emph{all} models $S$ of the given theory $T$.  
%\paragraph{Explain} explain why a given atomic formula is true / false in all models of $T$, or, in $T$ has no models, then explain the inconsistency.
%\paragraph{Determine Range}  determine the range of possible values for a given function term $f(\vec{t})$ given a theory $T$, i.e., the set of all values $v$ such that there exists at least one model $S$ of $T$ in which $f^S(\vec{t}^S) = v$. 
%\paragraph{Relevance} determine which symbols $\sigma$ are  relevant, in the sense that there exists a model $S$ of $T$ such that $S$ would no longer be a model if the value of $\sigma$ in $S$ were different.
%\paragraph{Logical entailment} verify whether a given statement $\phi$ is logically entailed by the theory $T$.

All of these were either already directly offered by IDP-Z3 (1-7) or trivial to implement (8).
This demonstrates that IDP-Z3 indeed provides the versatility that is required by the VERUS-LM framework.

% \item \textbf{Multiple choice}: proposes several different solutions to a given question. 
%\end{enumerate}
%\begin{algorithm}
%\caption{Solve Multiple Choice Question}
%\small
%\begin{algorithmic}[1]
%\Procedure{MultipleChoice}{$question$, $options$, $idp$}
    %\State $inference \gets \text{detect\_inference}(question)$
    %\If{$inference \in \{\textit{optimization}, \textit{range}\}$}
        %\State $result$, $goal$ $\gets$ $\text{execute}(question, inference, idp)$
        %\If{$result \in options$} 
        %\State \Return $result$
        %\Else
        %\State $inference \gets \textit{propagation}$
        %\EndIf
    %\EndIf
    %\State $idp \gets \text{add\_options}(idp, options)$
    %\State $pos\_options \gets \text{execute}(question, inference, idp)$
    %\If{$inference = \textit{optimization}$}
        %\If {$goal.max$} \Return $\max(pos\_options)$ 
        %\Else \hspace{0.15em} \Return $\min(pos\_options)$ 
        %\EndIf
    %\ElsIf{len($pos\_options$) = 1}
        %\State\Return $pos\_options[0]$
    %\Else
        %\State \Return $choose\_options(pos\_options)$
    %\EndIf
%\EndProcedure
%\end{algorithmic}
%\label{alg:multi}
%\end{algorithm}
Detecting the correct inference to be executed is crucial to ensure that questions are correctly answered. To detect the intended inference from the a question, we fine-tuned a BERT sentence classifier~\citep{sentencebert} on a custom dataset, which contains over 1000 natural language questions and their corresponding logical reasoning task. 

To address more complex nested questions that demand a sequence of logical reasoning steps, we developed a multi-step reasoning mode. Similar to Chain-of-Thought prompting, this mode first breaks the complex question into a series of steps. Each step is then treated as an independent reasoning task, after which the results are combined.

\subsection{Information Extraction}

Several of the forms of reasoning listed above require some additional information, in addition to a theory $T$. We distinguish two cases. % This includes the For instance, optimization also requires the term $t$ that must be minimized/maximized.  We de
%During the \textbf{information extraction} step, the VERUS-LM framework makes a distinction between extracting data and formulating additional constraints given the question.

\paragraph{Extracting data and goal detection}

Typically, in addition to a theory $T$, an interpretation is given for at least some of the symbols of $T$. For instance, the problem statement may be: ``Given that Ann is 16 and Brit is 32, who is eligible for insurance?'' Such information should be extracted and included as part of a structure on which the reasoning is performed (in this case, $age^S = \{ \text{Ann} \mapsto 16,\text{Brit} \mapsto 32\}$). In addition, some other reasoning tasks also require additional information of similar complexity, e.g., optimization requires that we know which term $t$ should be minimized / maximized.

This information extraction task is made significantly easier by the fact that, at this point, we already know the vocabulary.
% We can therefore use language model grammars, such as llama.cpp GNBF grammar~\citep{llamacpp}, to not only restrict the output of the model to correct syntax (e.g., json format), but to also force the output to adhere to an automatically generated KB-specific grammar, which ensures that only correctly typed interpretations are generated.
We can therefore use language model grammars, such as llama.cpp GNBF grammar~\citep{llamacpp}, to force the output to adhere to an automatically generated KB-specific grammar, ensuring that only correctly typed interpretations are generated.
% As we will show, this pruning of the output space allows us to achieves the same performance for this task with a small language model~(SLM) as with a larger LLM. Since such a SLM can run on a standard laptop, this may significantly reduce computational costs, as well as enhance data security, by avoiding the transfer of potentially sensitive information (e.g., the personal data of customers who apply for insurance) to external servers.
As we will show, this pruning of the output space allows us to achieves the same performance for this task with a small language model~(SLM) as with a larger LLM. Since such a SLM can run on a standard laptop, this may significantly reduce computational costs, as well as enhance data security, by avoiding the transfer of potentially sensitive information to external servers.

%As indicated by the laptop icon in Figure \ref{fig:VERUS-LM}, most of the inference phase—except for adding constraints—relies on lightweight models like the BERT sentence classifier and PHI 3.5 Mini-Instruct, a 3.8-billion-parameter language model. 

%When the question includes only data, i.e. (partial) information about a scenario, a language model is instructed to identify and translate relevant information into symbol assignments. This process relies on clear instructions and well-defined examples. 

%By using language model grammars, such as llama.cpp GNBF grammar~\citep{llamacpp}, we are able to restricts the output format of the model (e.g json format). 

%The typed nature of \fodot facilitates to go even a step further then just restricting the format of the output. By automatically generating KB-specific grammars, we can effectively restrict the models output space to a set of only valid symbol assignments for the given KB. 

\paragraph{Construct complex formulas}

%\benjamin{Het toevoegen van constraints is niet perse specifiek voor logical entailement, dit kan ook gebeuren voor andere reasoning tasks. vb what is the most amount of genres that can have both new and old cd's on sale. dan wordt er ook een symbol + constraint toegevoed en deze dan geoptimaliseerd, why is it not possible to have either X or Y on sale?}
%\simon{Heb de eerste zin wat aangepast om duidelijker te maken dat het niet entailment-specifiek is. Ok zo?}
In some case, such as for the logical entailment reasoning task, more complex additional information is required at inference time (e.g., the formula $\phi$ for which it should be checked whether $T\models \phi$). Constructing a complex formula $\phi$ from the natural language question is of similar complexity as constructing the knowledge base in the first phase. Therefore, SLMs are ill-suited for this task, and a larger LLM should be used instead. Similarly to the KB creation step, this LLM not only generates formulas but may also add new symbols to the vocabulary.

%\joost{Nieuw paragraafje:}
We currently do not attempt to automatically detect whether a specific question contains complex formulas or not, and thus requires an LLM, because we may wish to make this decision depending on the application. If we do not care about computational cost, we can just always invoke the LLM for optimal accuracy. Alternatively, we could base the decision whether to use an SLM or LLM purely on the detected reasoning task (e.g., Entailment is typically more likely to need an LLM), or we could even train a specific classifier to make the decision.

%If the question requires formulating additional constraints beyond symbol assignments, the expressiveness of \fodot makes predefined grammars less effective. For this reason

%The monotonicity of IDP-Z3 ensures that adding extra constraints or symbols does not invalidate the original KB. 

%Note that, any symbol assignment as described above, could also be represented as a set of additional constraints.

%\paragraph{Goal Detection}
%Certain reasoning tasks involve detecting a specific goal, such as identifying the numeric symbol for optimization (and determining if a maximum or minimum is looked for) or finding the symbol whose range must be evaluated. Similar to the data extraction step, KB-specific grammars are employed to constrain the output of small language models (SLMs) effectively. For each type of inference requiring a goal, a tailored prompt and grammar is designed to guide the model’s output.

%\paragraph{Local Inference Capability}
%This structured process and validation step enable KB Reasoning to ensure that user questions are accurately interpreted and addressed.

%We evaluate VERUS-LM on five widely-used benchmarks, and also introduce one of our own.
%We distinguish between single-reasoning and diverse-reasoning datasets.
%The former are datasets that focus exclusively on a single of reasoning task (e.g., deductive reasoning), while the latter have a more diverse set of questions.

\section{Experiments}

% \simon{Nog een korte reminder: overal ofwel dataset ofwel dataset schrijven. Ik denk dat het tweede de correcte vorm is?}\joost{Wikipedia zegt: A dataset (or dataset) is a collection of data. Ik zou zonder spatie doen, dat leest makkelijker.}

\subsection{Validation}
We first conduct a basic validation of VERUS-LM by checking (1) whether our use of a reasoning engine indeed leads to better perfomance on logical reasoning tasks than a stand-alone LLMs and (2) that the VERUS-LM pipeline is indeed able to correctly identify our eight different reasoning tasks from natural language text.  Since existing datasets typically cover only one or at most a few reasoning tasks, we constructed our own dataset called \emph{DivLR}    in which all eight reasoning tasks are present. It consists of 115 questions covering six domains:  investment \textbf{S}trategies, \textbf{C}OVID restrictions, water \textbf{I}rrigation needs, a \textbf{H}andyman and two variants in the Body Mass Index (BMI) domain. The reason for having two BMI variants is that information about BMI (i.e., how to calculate it, the different risk levels associated to BMI values, etc.) probably occurs in LLM training data. In addition to our  \textbf{B}MI domain which contains the well-known definition, we also defined a \textbf{B}\textbf{*} domain in which novel formulas and ranges for BMI were made up, similar to the experiment in~\citep{Goossens2023a}.

We compare VERUS-LM to baseline language models on this dataset. We compare with both an SLM and an LLM, and consider two versions of our pipeline: one in which our information extraction is done by an SLM (V-SLM) and one in which this is done by an LLM (V-SLM). the LLM is Gemini 1.5 Pro~\citep{team2023gemini} and the SLM is a quantised version of Phi 3.5~\citep{phi3}, a 3.8-billion-parameter language model whose inference cost is about 10,000 times lower than large models like GPT-4. All experiments were performed on a Mac M2 Pro CPU with 32GB of RAM.

\begin{table}
    \centering
    \caption{Results of VERUS-LM with SLM and LLM for information extraction and SLM and LLM separately as baseline. SLM = Phi 3.5 mini instruct, LLM = Gemini 1.5 Pro}
    \label{table:custom}
    \begin{tabular}{lrrrrrrr}
        \toprule
        System & C & S & H & I & B & B* & Avg \\
        \midrule
        V-SLM & \textbf{94.7} & 86.4 & 86.7 & 86.4 & 61.1 & 66.7 & 80.3 \\ 
        V-LLM & 89.5 & \textbf{100} & \textbf{93.3} & \textbf{95.5} & \textbf{88.9} & \textbf{83.3} & \textbf{91.8} \\ 
        SLM & 47.4 & 50 & 53.3 & 36.4 & 38.9 & 11.1 & 39.5 \\ 
        LLM & 73.7 & 86.4 & 60 & 63.6 & 66.7 & 50 & 66.7 \\
        \bottomrule
    \end{tabular}
\end{table}

Table \ref{table:custom} shows that both V-SLM and V-LLM consistently outperform both the SLM and the LLM on their own. However, V-SLM performs significantly worse on the two BMI domains than on the other domain, and also significantly worse than V-LLM on these domains.  A deeper analysis shows that the SLM typically refuses to simply extract the information that the reasoning engine needs, and instead tries to perform BMI-calculations itself. We conjecture that this is due to the BMI-related examples that are likely part of the data on which the SLM was trained.

A similar but less striking issue can be seen when comparing the results of the LLM on both the \textbf{B} and \textbf{B}\textbf{*} domain: the observed drop in accuracy (66.7$\rightarrow$50) can be explained by the fact that the LLM has memorised the correct definition of BMI from its training data, and that is  unable reason with a different definition, when one is explicitly provided. This suggests that, more generally, if an LLM at some point learns out-dated facts, it is hard to correct these in the prompt, requiring retraining the network instead.  We see that same issue is still somewhat observable in V-LLM (88.9$\rightarrow$83.3), but here the reasoning engine pipeline substantially reduces the effect.

\begin{table}
    \centering
    \caption{Distribution of different reasoning tasks in custom datasets, detection accuracy of the BERT classifier and execution accuracy of VERUS-LM with an SLM (V-SLM) and LLM (V-LLM) for information extraction.}
    \label{table:custom-task}
    \begin{tabular}{lcccc}
    \toprule
        \multirow{2}{*}{\parbox{3em}{\vspace{0.4em}Reasoning Task}}& \multirow{2}{*}{\parbox{3em}{\vspace{0.5em}\makecell{\%}}} & \multirow{2}{*}{\parbox{3em}{\vspace{0.3em}\makecell{Detect\\Acc}}} & \multicolumn{2}{c}{Exec\_Acc} \\
        \cmidrule(lr){4-5}
         &  &  & V-SLM & V-LLM \\
    \midrule
        Model Expansion& 1.8 & 100 & 100 & 100 \\ 
        Satisfiability & 9.6 & 81.8 & 100 & 90.9 \\
        Optimization& 26.3 & 90.0 & 93.3 & 96.7 \\
        Propagation& 28.1 & 76.5 & 68.6 & 91.2 \\
        Explain& 8.8 & 100 & 80.0 & 90.0 \\ 
        Determine Range& 14.0 & 92.9 & 100 & 85.7 \\ 
        Relevance& 1.8 & 100 & 100 & 100 \\ 
        Entailment& 9.6 & 100 & 27.3 & 90.9 \\
    \bottomrule
    \end{tabular}
\end{table}

Table \ref{table:custom-task} confirms that our BERT classifier identifies the correct logical reasoning task with an average accuracy of 92.6\%, performing consistently well on all tasks (with a small dip for Propagation).  For the subsequent information extraction, we see that, for most of the reasoning tasks, V-LLM and S-SLM perform about equally well.  The biggest difference is seen for Entailement. This is the only reasoning task in our DivLR dataset for which complex formulas need to be constructed at inference time, which V-SLM predictably struggles with (accuracy 29.6\%). For the Propagation task, we also observe a difference in accuracy (-25.6\%) between V-SLM and V-LLM. This is primarily due V-SLM struggling with the BMI domain, as already discussed.

Overall, we conclude that the proposed pipeline makes sense: (1) introducing a logical reasoning engine into the pipeline indeed leads to better performance than using only a language model; (2) different logical reasoning tasks can indeed by successfully identified by a simple BERT classifier; (3) all of the necessary information for the reasoning tasks can indeed be extracted with a small language model, as long as no complex formulas need to be constructed, and with the caveat that an SLM might be less robust when handling concepts that already occurred frequently in its training data.

In our following experiments, we have always used V-SLM for benchmark datasets that do not require the construction of complex formulas at inference time and V-LLM otherwise.

\subsection{Comparison to state-of-the-art}

\begin{table*}[!t]
    \small
    \centering
    \caption{Overview of datasets used for comparison with state-of-the-art}
    \label{table:datasets}
    \begin{tabular}{lcccccc}
    \toprule
        \makecell{Dataset} & \makecell{Example} & \makecell{For inference} & \makecell{\# Options} &\makecell{Reasoning} & \makecell{\#}\\ 
        \midrule
        \makecell{PrOntoQA\\\citep{prontoQA}} & \makecell{``Stella is not hot''} & Data & \makecell{2: $\top$, $\bot$}&\makecell{Satisfiability} & 500 \\
        %\hdashline
        \midrule
        \makecell{ProofWriter\\\citep{proofwriter}} & \makecell{``The rabbit visits the cat''} & Data & \makecell{3: $\top$,$\bot$, ?} &\makecell{Propagation} & 600 \\
        \midrule
        %\hdashline
        \makecell{FOLIO\\\citep{folio}} & \makecell{``Marvin is neither a\\human nor from Mars''} & Formulas & \makecell{3: $\top$, $\bot$, ?}  &\makecell{Logical\\entailment} & 204 \\ 
        %\hdashline \makecell{LogicalDeduction\\\citep{logicaldeduction}} & \makecell{``A)Eve finished third,\\B) Max finished third, $\dots$''} & \makecell{Only Data} & \makecell{3-5-7} &\makecell{Satisfiability} & 300 \\
        \midrule
        \makecell{LogicalDeduction\\\cite{logicaldeduction}} & \makecell{A)"Eve finished third.",\\B) "Max finished third", $\ldots$} & \makecell{Only Data} & \makecell{3-5-7} &\makecell{Satisfiability} & 300 \\
        %\hdashline
        \midrule
        \makecell{AR-LSAT\\\citep{arlsat}} & \makecell{``What is the max \#CDs on \\sale? A) two, B) six, \ldots''} & Formulas & \makecell{5} & \makecell{Multiple} & \makecell{231} \\
    \bottomrule
    \end{tabular}
\end{table*}
\begin{table*}[!t]
    \centering
    \caption{Comparison of the results of the neurosymbolic system. $^*$Results from original paper. $^\dag$Results from \protect\citep{symbocot}.
        % Results for Logic-LM, SymboCOT and LINC come from their respective papers \protect\citep{logiclm,symbocot,linc}, and the LLM baselines originate from \protect\citep{symbocot} as it is the most recently published paper.
    The results for Logic-LM are those with self refinements included, but without the LLM fallback in case of recurring syntax error. %The best results per dataset are put in bold, the second best results are italicized.}%\joost{Ik zou de 2nd best ``italicized'' maken ipv. underlined, want door de stippellijnen zijn de onderlijningen moeilijk te zien.}}
    }
    \label{table:benchmark}
    \begin{tabular}{lccccccc}
        \toprule
        System & ProntoQA & Proofwriter & Folio & \makecell{Logical\\Deduction} & AR-LSAT & Average & ST Dev \\
        \midrule
        VERUS-LM & 95.8 & \textit{93.83} & \textit{78.43} & \textit{88.67} & \textbf{68.36} & \textbf{85.02} & \textbf{11.49}\\
        %\hdashline
        Logic-LM$^*$ & 83.20 & 78.80 & 68.55 & 87.63 & 23.40 & 68.32 & 26.09 \\
        LINC$^*$ & x & \textbf{98} & 72.50 & x & x & x & x \\
        %\hdashline
        SymbCoT$^*$ & \textbf{99.60} & 82.50 & \textbf{83.33} & \textbf{93.00} & \textit{43.91} & \textit{80.47} & 21.63 \\
        %GPT3.5 & 51.80 & 36.16 & 54.60 & 41.33 & 22.51 & 41.28 & \textit{11.55} \\
        %GPT3.5 - CoT & 83.00 & 48.33 & 57.84 & 48.33 & 22.51 & 52.00 & 19.45 \\
        GPT4$^\dag$ & 77.40 & 52.67 & 69.11 & 71.33 & 33.33 & 60.77 & \textit{17.86} \\
        GPT4 - CoT$^\dag$ & \textit{98.79} & 68.11 & 70.58 & 75.25 & 35.06 & 69.56 & 22.80 \\
        \bottomrule
    \end{tabular}
    %\caption{Accuracy of VERUS-LM, 3 neurosymbolic approaches (LogicLM \citep{logiclm}, SymboCOT \citep{symbocot}, LINC \citep{linc}) and baseline LLMs using standard and chain-of-thought prompting on five reasoning datasets. The best results per dataset are put in bold, the second best results are underlined.}
\end{table*}

Having established that the VERUS-LM pipeline makes sense, we now compare it to the state-of-the-art on a number of standard benchmarks, summarized in Table \ref{table:datasets}. PrOntoQA, ProofWriter, FOLIO, and LogicalDeduction each target a specific reasoning task, while the challenging AR-LSAT dataset spans a broader range of reasoning tasks.

We compare VERUS-LM against three LLM-based approaches and two neurosymbolic approaches for logical reasoning: (1) GPT-4 answering directly, (2) GPT-4 with Chain-of-Thought, (3) 
Linc~\citep{linc} (only FOLIO and ProofWriter), (4) Logic-LM~\citep{logiclm} and (5) SymbCot~\citep{symbocot}.
As shown in Table~\ref{table:benchmark}, all systems achieve a roughly similar performance on the first four benchmarks, with VERUS-LM scoring at most 5\% worse than the best system on each, without relying on \textbf{dataset-specific prompts}.
For the challenging AR-LSAT benchmark, VERUS-LM  significantly outperforms all other methods, doing about \textbf{25\% better} than the second best system. We believe that this is due to (1) the \textbf{generality} of our approach, which does not use task-specific examples for in-context learning; (2) \textbf{\fodot's expressiveness} which allows for, e.g., straightforward representations of aggregates, such as ``At least three CDs are on sale'': $\#\{ c \in \textit{CD}: on\_sale(c)\} \geq 3$; (3) our support for \textbf{different forms or reasoning}, four of which were detected by the classifier and executed, as shown in Table \ref{table:distribution}.

%In addition to achieving competitive results for these dedicated datasets, VERUS-LM significantly outperforms the other approaches by a wide margin on the AR-LSAT dataset. %Several factors contribute to the superior performance of VERUS-LM over other neurosymbolic methods:
%We feel that several factors contribute to this jump:

%\simon{Iets over de stdev? Lijkt me zeker waardevol!} \benjamin{ja maar die st dev wordt natuurlijk vooral beinvloed door de lage AR-LSAT score wat hier nog niet echt van toepassing is}
%\paragraph{AR-LSAT}
\begin{table}
    \centering
    \caption{The distribution of reasoning tasks detected in AR-LSAT and VERUS-LM's accuracy}
    \label{table:distribution}
    \begin{tabular}{lcc}
    \toprule
    Reasoning task & Distribution (\%) & Exe\_Acc (\%) \\
    \midrule
    Optimization & 6.06 & 71.43 \\ 
    Determine range & 16.88 & 79.49 \\ 
    Entailment & 45.89 & 66.04 \\ 
    Satisfiability & 31.17 & 69.01 \\
    \bottomrule
    \end{tabular}
\end{table}
%\begin{figure}
    %\centering
    %\includegraphics[width=1\linewidth]{img/ar-lsat.png}
    %\caption{The distribution of reasoning tasks detected in AR-LSAT dataset}
    %\label{fig:enter-label}
%\end{figure}

%\textbf{1. In-context learning is not robust.} While it is effective datasets with uniform knowledge structures, in-context learning lacks generality. This limitation becomes evident in the results of the AR-LSAT dataset which encompasses a diverse range of domains and knowledge types. VERUS-LM's more robust workflow, with general examples for demonstration and a semantic additional refinement step, proves more adaptable to such diversity.

%\simon{Heb hier even de enumeratie herschreven, verander maar terug indien gewenst}

%\joost{Voor mij hoeft dit er niet echt bij. Het zou nuttig zijn om ook een tabelletje te hebben met bv.~de gemiddelde computational cost per vraag voor de verschillende systemen, maar als we dat niet hebben, heeft het ook niet zoveel zin om hier een beetje te handwaven over drie van de systemen en niets te zeggen over te rest. achieves competitive results across the four reasoning-task datasets.  While SymbCoT slightly outperforms VERUS-LM, it incurs higher computational costs due to extended reasoning chains. Compared to Logic-LM, our approach adds overhead by separating the symbol extraction from constraint formulation, and adding a semantic refinement step but this is greatly compensated through KB reuse, where a single KB sometimes answers up to seven questions.}

%\paragraph{Self-Refinement Impact}
\subsection{Effect of Self-Refinement}
\begin{table}
    \centering
    \caption{Execution Rate and Accuracy of VERUS-LM in different refinements scenario's: \textbf{No} Refinements, with \textbf{Syntax} Refinements and with \textbf{Both} Syntax and Semantic Refinements}
    \label{table:ref-analysis}
    \footnotesize
    {\tablefont\begin{tabular}{@{\extracolsep{\fill}}lclll}
        \toprule
        ~ & ~~~~Refs~~~~ & ~~~~Exe\_Rate~~~~ & ~~~~Exe\_Acc~~~~ & ~~~~Total\_Acc \\
        \midrule
         & No & 74 & 97 & 71.8 \\
        PrOntoQA & Syntax & 88.2 \textcolor{Green}{$\uparrow$ 14.2} & 97.3 \textcolor{Green}{$\uparrow$ 0.3}  & 86.8 \textcolor{Green}{$\uparrow$ 15} \\
        & Both & 98.2 \textcolor{Green}{$\uparrow$ 10} & 97.6 \textcolor{Green}{$\uparrow$ 0.3}& 95.8 \textcolor{Green}{$\uparrow$ 9} \\
        %\hdashline
        \midrule
        & No & 90 & 95.7 & 86.2 \\ 
        ProofWriter & Syntax& 92.3 \textcolor{Green}{$\uparrow$ 2.3} & 95.7 \textcolor{gray}{$=$} & 88.3 \textcolor{Green}{$\uparrow$ 2.1} \\ 
        & Both & 99 \textcolor{Green}{$\uparrow$ 6.7} & 94.8 \textcolor{Red}{$\downarrow$ 0.9} & 93.8 \textcolor{Green}{$\uparrow$ 5.5}  \\
        \midrule
        %\hdashline
         & No & 71.6 & 80.8 & 57.8 \\ 
        FOLIO & Syntax& 89.2 \textcolor{Green}{$\uparrow$ 17.6} & 80.6 \textcolor{Red}{$\downarrow$ 0.2}& 74 \textcolor{Green}{$\uparrow$ 16.2 }\\ 
        & Both & 100 \textcolor{Green}{$\uparrow$ 10.8} & 78.4 \textcolor{Red}{$\downarrow$ 2.2}& 78.4 \textcolor{Green}{$\uparrow$ 4.4} \\
        \midrule
        %\hdashline
        \multirow{3}{*}{\makecell[c]{Logical\\ Deduction}} 
    & No       & 93.3 & 89.6 & 83.6 \\ 
    & Syntax      & 93.7 \textcolor{Green}{$\uparrow$ 0.4} & 89.7 \textcolor{Green}{$\uparrow$ 0.1} & 84 \textcolor{Green}{$\uparrow$ 0.4} \\ 
    & Both  & 99.3 \textcolor{Green}{$\uparrow$ 6.3} & 89.3 \textcolor{Red}{$\downarrow$ 0.4} & 88.7 \textcolor{Green}{$\uparrow$ 4.7} \\
        \midrule
        %\hdashline
          & No & 60.2 & 84.3 & 50.8  \\ 
        AR-LSAT & Syntax & 81.8 \textcolor{Green}{$\uparrow$ 21.6} & 78.2 \textcolor{Red}{$\downarrow$ 6.2} & 64 \textcolor{Green}{$\uparrow$ 13.2}\\ 
        & Both & 98.7 \textcolor{Green}{$\uparrow$ 16.9} & 69.3 \textcolor{Red}{$\downarrow$ 8.9} & 68.4 \textcolor{Green}{$\uparrow$ 4.4} \\
        \bottomrule
    \end{tabular}}
\end{table}

Table \ref{table:ref-analysis} presents the results for each dataset of: (1) VERUS-LM without performing any refinements after the initial KB creation; (2) VERUS-LM with only syntactic refinement; (3) complete VERUS-LM with also semantic refinements. In each case, we report the \emph{execution rate}, which is the percentage of cases in which the knowledge base was syntactically correct and satisfiable. We also report the \emph{execution accuracy}, which is the percentage of the those for which the reasoning engine returned a correct result. We also report the product of the two as ``total accuracy''.

The syntactic refinement increases the execution rate by 11.2\% on average. The execution accuracy  stays roughly the same, showing that KBs that had to be syntactically corrected are about as likely to correctly represent the domain knowledge as KBs that were initially already syntactically correct. When we then also include our semantic refinement, there is an additional increase in execution rate of on average  10\%. %\simon{We moeten die averages zeker nog eens herberekenen aleer in te dienen}.\benjamin{done} 
This concerns cases in which the original KB was unsatisfiable, e.g., because of contradictory formulas or mistakes in the typing of a function or constant. Such errors typically arise only when the natural language description is difficult to understand, which also makes them  hard to rectify in a correct way, as evidenced by the drop in execution accuracy when adding the semantic refinement. However, the total accuracy still markedly increases, showing that the semantic refinement step is indeed useful.

\section{Limitations}
Though VERUS-LM expands on the state of the art, it faces some limitations.
Like the earlier papers in literature, VERUS-LM itself does not have a method for validating formula correctness.
This is an inherent problem in this type of neuro-symbolic AI in general, as natural language can be ambiguous, and LLMs can always produce errors.
To address this limitation, we are looking into bringing a domain expert in the loop without prior logic experience to validate the output~\cite{Voet2025}.

Additionally, the expressiveness of \fodot has limitations, making it unsuitable  for, e.g., representing higher-order logic or for reasoning over OWA.
Though VERUS-LM's simulated OWA performed well on the datasets discussed in this work, it remains an approximation and thus will not be effective for all situations.
Similarly, our pipeline's semantic refinement step is also not always applicable, as there could be situations (such as reasoning over fake news) where we \textit{can} expect unsatisfiability.
Due to VERUS-LM's relience on a logical reasoning engine, it also inherits some of its limitations, such as SAT problems in very large domains. %\simon{Dit is niet per se inherent aan IDP-Z3 -- alle logical solvers hebben dit probleem}
Furthermore, though the KB-creation of VERUS-LM is based on a generic prompting pipeline, its accuracy in entirely different logical paradigms remains a uncertain.
% \benjamin{misschien is het goed om hier ook kort te algemene limitations van deze neurosymbolisch systemene aan te kaarten rondom te certainty of the translation}
% \simon{Plus vermelden dat de semantic refinement check niet altijd wenselijk is}
%\simon{Nog eentje: de benchmark datasets zijn overwegend kleine problemen, we hebben nog geen idee of VERUS-LM even goed kan scoren op problemen van realistischere grootte.}
%\benjamin{Gohja, de custom domain zijn best wel groot hoor}
%\simon{Ja, maar niet zo groot als b.v. de SG use case of de Adhesive Selector. ;-)}
%\benjamin{Ja maar, om dat nu effectief als een limitaion aan te duiden dunno + staat da nie ook een beetje in die SAT problemen}
%\simon{Niet per se -- zie het als drie grootte-ordes: ``dummy problem'', ``medium realistic size'', ``very large domain''. We weten nu al dat de eerste lukt, en we vermelden dat de derde niet gaat lukken door limitaties van de interne solver. Dat tweede, daarentegen, lukt misschien wel, maar misschien ook niet. Nochtans is dat het interessante: met enkel dummy problemen op te lossen doen we weinig nuttigs.}

\section{Conclusion}
This paper introduced VERUS-LM, a versatile neurosymbolic framework that integrates language models with symbolic reasoning capabilities. Its two-phased approach separates Knowledge Base creation from a separate inference phase, in which multiple different reasoning tasks can be performed on the same knowledge base. This approach has the inherent advantage that multiple questions about the same domain can be answered in a computationally efficient way, by reusing the same knowledge base. Another improvement made by VERUS-LM is that we  extend the state-of-the-art syntactic refinement step with a semantic refinement step, based on a satisfiability check. Our experimental analysis shows that this indeed improves results. VERUS-LM significantly outperforms the state-of-the-art on the challenging and diverse AR-LSAT dataset, wile remaining competitive on simpler benchmarks. %Furthermore, we aim to enhance the quality and robustness of the KB creation pipeline by fine-tuning a language model for generating \fodot programs, and validating our framework on larger, real-world use cases.

%used in the statemploys a generic KB creation pipeline with a semantic refinement step, ensuring adaptability to diverse domains without relying on in-context learning, in contrast to state of the art methods. This design achieves competitive results on dedicated benchmarks and outperforms the SotA on the challenging and diverse AR-LSAT dataset. Additionally, VERUS-LM’s support for a wide range of reasoning tasks demonstrates its ability to handle diverse question types, as evidenced by its performance on our DivLR dataset.

\bibliographystyle{eptcs}
\bibliography{newbib}

\end{document}